\documentclass{article}

\usepackage{arxiv}

\usepackage[utf8]{inputenc}
\usepackage[T1]{fontenc}
\usepackage{hyperref}
\usepackage{url}
\usepackage{booktabs}
\usepackage{amsmath}
\usepackage{amsthm}
\usepackage{amsfonts}
\usepackage{amssymb}
\usepackage{nicefrac}
\usepackage{microtype}
\usepackage{graphicx}
\usepackage{caption}
\usepackage{subcaption}
\usepackage{multirow}
\usepackage{threeparttable}
\usepackage{float}
\usepackage{xcolor}
\usepackage{doi}
\usepackage[titletoc,title]{appendix}

\usepackage{frcursive} 

\newcommand{\Sigv}{\mbox{\boldmath$\Sigma$}}

\newcommand{\mv}{\mbox{\boldmath$\mu$}}

\newcommand{\xv}{\mbox{\boldmath$x$}}

\newcommand{\wv}{\mbox{\boldmath$w$}}

\newcommand{\pmc}{P_{\rm mc}}

\title{PHM-ensemble: Scalable and Stable Probabilistic Multiresolution Clustering via Ensemble Density Estimation}

\renewcommand{\shorttitle}{PHM-ensemble: Scalable and Stable Multiresolution Clustering}
\renewcommand{\headeright}{}
\renewcommand{\undertitle}{}

\author{
  Ali Turfah \\
  Department of Biostatistics\\
  University of Michigan\\
  Ann Arbor, MI 48109, USA \\
  \texttt{aturfah@umich.edu} \\
  \And
  Xiaoquan Wen \\
  Department of Biostatistics\\
  University of Michigan\\
  Ann Arbor, MI 48109, USA \\
  \texttt{xwen@umich.edu} \\
}

\date{}

\hypersetup{
pdftitle={PHM-ensemble: Scalable and Stable Probabilistic Multiresolution Clustering via Ensemble Density Estimation},
pdfsubject={stat.ML, stat.ME, q-bio.QM},
pdfauthor={Ali Turfah, Xiaoquan Wen},
pdfkeywords={Unsupervised learning, Cluster analysis, Multiresolution clustering, Gaussian mixture models, Density Estimation, Representation Learning, Ensemble learning, Single-cell RNA-seq},
}

\begin{document}

\maketitle

\begin{abstract}
Clustering is a foundational tool for exploratory data analysis across the sciences.
We argue that clustering methods should be evaluated along four dimensions: fidelity to the data's underlying structure, interpretability of the recovered groups, stability under data perturbations and algorithmic randomness, and scalability to modern scientific datasets.
Yet existing methods rarely satisfy all four criteria.
To address this gap, we introduce PHM-ensemble, a general probabilistic framework for multiresolution clustering that estimates the underlying data density by pooling mixture models from an ensemble of fits and subsequently constructs a clustering hierarchy through hierarchical merging of the mixture components.
On simulated datasets and the MNIST benchmark, PHM-ensemble matches or outperforms widely used clustering algorithms in agreement with ground truth, produces more stable results across runs, and reveals complex underlying structures.
Applied to 1.82 million peripheral blood cells from the Human Immune Health Atlas, PHM-ensemble achieves higher agreement with expert-curated major cell-type annotations than the Leiden clustering used in the original study (adjusted Rand index 0.965 versus 0.789), while requiring only half the computing time.
Beyond identifying major immune lineages, it reveals biologically coherent relationships among cell types and resolves finer structure within individual lineages, uncovering subtypes and cellular states consistent with known biology.
\end{abstract}


\section{Introduction} 
 
Clustering is a widely used computational tool for unsupervised data analysis, allowing researchers to uncover structure in data without prior knowledge of the underlying group labels. It is particularly useful for exploratory data analysis and has seen widespread use across a broad range of scientific disciplines.
In genomics, clustering is routinely used to identify cell types from single-cell gene expression data; in astronomy, to classify galaxies and stars based on photometric and spectral measurements; in economics, to segment markets and consumer populations; and in social science, to identify communities within social networks. 
However, as datasets in these fields continue to grow in size and complexity, existing clustering methods face increasing challenges.
Meeting the demands of contemporary scientific applications requires new clustering methods that improve fidelity, interpretability, stability, and scalability simultaneously.

Fidelity and interpretability are closely related, concerning the quality and representation of clustering results, respectively. 
We use fidelity to refer to the extent to which the output reflects the data's true underlying structure, i.e., whether the clustering faithfully captures the patterns in the data.
Interpretability, by contrast, concerns whether the recovered cluster structure can be understood and meaningfully interpreted by practitioners: are the clusters clearly distinct, do they correspond to known covariates or domain knowledge, and can the clustering results be readily explained? 
While fidelity is inherently important, interpretability becomes critical when ground truth is unavailable, as is often the case in unsupervised learning.     
Many existing clustering methods represent the underlying structure through a single flat partition of the data. 
However, in many scientific applications, the underlying group structure arises from continuous differentiation processes and is inherently hierarchical. 
Collapsing such complex structure into a single flat partition can compromise both fidelity and interpretability. 
A multiresolution representation addresses these limitations by preserving structure across multiple scales, allowing practitioners to select the granularity appropriate to a particular scientific question and revealing relationships among groups and subgroups across resolutions.

Stability was originally defined as the robustness of a deterministic algorithm's output to perturbations in the input data \cite{bousquet2002stability}, and was later extended to randomized algorithms \cite{elisseeff2005stability} to account for variation arising from the algorithm's own stochasticity. 
Both sources of variability are directly relevant to clustering.
Stability with respect to data perturbations has long been recognized as an important validation criterion \cite{benhur2002stability, lange2004stability, vonluxburg2010stability}: if the recovered structure is genuine rather than an artifact of a particular sample, it should be reproducible across different, equally valid samples. 
At the same time, many clustering procedures involve randomized components that introduce additional variability. 
For example, $k$-means and the EM algorithm for fitting Gaussian mixture models (GMMs) can produce different results on the same dataset, as different random initializations may lead to different local optima. 
Substantial instability implies that the recovered structure may reflect sampling variation or algorithmic randomness rather than genuine patterns in the data, hence undermining both fidelity and interpretability.  
Therefore, we consider stability an essential requirement for reliable clustering, particularly in scientific applications.

Modern scientific datasets, with rapidly increasing sample sizes and complexity, have outgrown the computational capabilities of many existing clustering methods.
A single-cell atlas can contain tens of millions of cells \cite{gong2024longitudinal}, while an astronomical survey can catalog hundreds of millions of objects \cite{abazajian2009seventh}. 
Many clustering methods with strong statistical foundations, including classical hierarchical clustering and Bayesian nonparametric approaches, scale poorly and can become computationally prohibitive for large datasets. More fundamentally, a clustering method has limited practical value if it cannot handle the scale of data routinely encountered in modern scientific applications, regardless of its fidelity, interpretability, or stability.
Thus, scalability is a practical prerequisite for realizing the benefits of fidelity, interpretability, and stability in modern scientific applications.

Motivated by the need for fidelity, interpretability, stability, and scalability, we propose a new and general computational framework for probabilistic multiresolution clustering. 
The proposed framework integrates ideas from statistical inference and representation learning.
It departs from the traditional view of clustering as a partitioning problem defined directly on the observed samples. 
Instead, we focus on learning the underlying data-generating distribution and interpreting its structure by constructing a cluster hierarchy. 
This distribution-focused perspective lets us leverage established machine learning techniques, including density estimation and ensemble learning, to improve the fidelity and stability of clustering results while reducing the computational burden of large-scale datasets.
Extending our previous work on the $\pmc$ hierarchical merging (PHM) algorithm \cite{wen2026probabilistic}, we propose constructing a cluster hierarchy directly from the estimated distribution, providing a multiresolution representation that reveals structure across scales rather than committing to a single resolution in advance and thereby improving interpretability.

The remainder of this paper is organized as follows. 
Section 2 presents the rationale and methodology of the proposed clustering framework, together with an analysis of its computational complexity. 
Section 3 evaluates the proposed method using simulated and real benchmark datasets, including MNIST and an atlas-scale single-cell RNA-seq dataset. 
Section 4 concludes by discussing the main findings and future directions.


\section{Methods}

\subsection{Clustering as Probabilistic Representation Learning}

We view clustering as a representation learning task \cite{torralba2024foundations}. 
In this setting, the representation of interest is not a learned embedding or feature space, but rather the data-generating distribution underlying the observed data. 
Specifically, the goal is to estimate this distribution and interpret it through cluster structures.

Let $\xv \in \mathbb{R}^d$ denote a $d$-dimensional observation generated from an unknown distribution $P(\xv)$. 
We aim to learn this distribution from the observed data using a Gaussian mixture model (GMM). 
In the limit of an unbounded number of components, a GMM is a universal approximator of densities and can approximate $P(\xv)$ arbitrarily well \cite{goodfellow2016deep,genovese2000rates}; in practice, however, we work with a finite, data-driven approximation.
We denote the learned distribution by $\hat P(\xv)$ and assume that it consists of $K$ mixture components:
\begin{equation}
\hat P(\xv) = \sum_{k=1}^K \alpha_k \, \mathcal{N}(\xv; \mv_k, \Sigv_k),
\end{equation}
where $\alpha_k$ is the mixing proportion of the $k$th component, and $\mathcal{N}(\xv; \mv_k, \Sigv_k)$ denotes a multivariate Gaussian density with estimated mean vector $\mv_k$ and covariance matrix $\Sigv_k$.

Existing GMM-based clustering methods typically interpret mixture components directly as clusters \cite{fraley2002model}. 
However, the model-based clustering literature has long recognized that this one-to-one correspondence need not hold: a single Gaussian component is often too restrictive to represent a non-Gaussian subpopulation, and model-selection criteria such as BIC, which balance goodness of fit against model complexity, may select more components than meaningful clusters \cite{baudry2010combining,hennig2010merging}.
We adopt this more flexible view, modeling the overall data distribution as a Gaussian mixture in which each latent cluster may itself comprise multiple Gaussian components.
Under this formulation, clustering becomes the task of identifying and grouping inseparable or highly overlapping mixture components within the estimated distribution $\hat P(\xv)$.

This perspective naturally leads to a multiresolution view of clustering, which already underlies existing approaches that combine mixture components to represent meaningful clusters \cite{baudry2010combining,hennig2010merging}. 
More specifically, any finite mixture distribution admits a hierarchical representation obtained by recursively aggregating its atomic mixture components.
This follows directly from the associativity of convex combinations, which lets subsets of components be represented as ``macro-densities'' and subsequently grouped at higher levels of abstraction, without re-estimating densities at any stage of aggregation.
The resulting hierarchy is a nested representation of the population across multiple resolutions, ranging from the global distribution at the root (i.e., a single cluster) to increasingly localized subpopulations towards the leaves, while preserving the underlying probability distribution. 
However, such a hierarchy is generally not unique, as different aggregation strategies can produce substantially different cluster structures. 
Consequently, the central challenge in this framework is to quantify the overlap between mixture components and decide the order in which they should be aggregated. 
Although entropy- and misclassification-based methods have been proposed for merging mixture components \cite{baudry2010combining,hennig2010merging}, they are computationally intensive and do not scale well to large datasets.

Taken together, our proposed computational framework casts clustering not as an algorithmic partitioning problem defined directly on observed samples, but as a statistical inference problem of estimating the underlying data-generating distribution and identifying natural groupings within it.
Specifically, we decompose clustering into two sequential stages: (i) estimating the underlying probability distribution from the observed data, and (ii) constructing a cluster hierarchy that reveals natural groupings in the learned distribution across multiple resolutions. 
This decomposition allows us to address the challenges of large-scale data clustering separately at each stage, using computational techniques tailored to the task.
Importantly, the proposed framework does not abandon existing partition-based clustering methods. Instead, it can integrate these methods as an inductive bias to improve density estimation, as detailed in Section~\ref{sec:density-estimation}.

\subsection{Ensemble Density Estimation} \label{sec:density-estimation}

The Expectation-Maximization (EM) algorithm has been widely used to fit GMMs for density estimation \cite{dempster1977maximum}, including the popular \texttt{mclust} package \cite{mclust}.
However, when applied to large-scale datasets, standard EM faces several notable challenges.
First, GMM estimation via EM is sensitive to initialization, susceptible to local optima, and dependent on the choice of the number of components $K$ \cite{jin2016local, balakrishnan2017statistical, fraley2002model}.
These sensitivities can lead to unstable density estimates and, within our proposed framework, subsequently to degenerate clustering results.
Second, the per-iteration computational complexity of EM for fitting a $K$-component GMM to $N$ samples in $\mathbb{R}^d$ is $O(NKd^2)$, which poses substantial scalability challenges for large datasets.

To address these issues, we propose adopting an ensemble learning strategy: rather than relying on a single fitted GMM, the final density estimate is constructed by aggregating estimates across a collection of GMM fits.
Specifically, we denote the ensemble density estimate by
\begin{equation}
\hat P_e(\xv) = \sum_{m=1}^M w_m \hat P_m (\xv),
\end{equation}
where
\begin{equation*}
\hat P_m (\xv) = \sum_{k=1}^{K_m} \alpha_{mk} \, \mathcal{N} (\xv ; \mv_{mk}, \Sigv_{mk}).
\end{equation*}
The ensemble weight vector $\wv = (w_1,\ldots,w_M)$ lies on the probability simplex and represents the relative importance of the corresponding GMM fits in the collection.
In this paper, we focus primarily on a simple scheme that assigns equal weights to all GMM fits when computing the final density estimate, setting $w_m = 1/M$ for $m=1,\ldots, M$. Our software also supports other more general weighting schemes (Section 1.3 of the Supplementary Notes).
Each replicate density estimate, $\hat P_m(\xv)$, is obtained by applying the standard EM algorithm for Gaussian mixtures to an independently constructed subsample and initialization. Specifically, for each replicate, we first subsample the data, then construct a partition-guided initialization on the subsample, and finally fit a GMM by EM.
The resulting ensemble therefore captures two distinct sources of variability: subsampling perturbs the observed data while reducing the computational burden of fitting EM to the full dataset, whereas initialization diversity mitigates EM's sensitivity to local optima in the non-convex likelihood surface \cite{jin2016local}.

Subsampling is controlled by a user-specified proportion $\rho \in (0,1]$.
For each replicate $m$, we draw $\lfloor \rho N \rfloor$ observations {\em without}
replacement from the full dataset of size $N$, independently across
replicates.
This scheme, known as subsample aggregating (i.e., subagging), offers a computationally cheaper alternative to bootstrap resampling with replacement at the full sample size $N$ (i.e., bagging)  \cite{buhlmann2002analyzing}.
While the subsampling with multiple replicates does not necessarily reduce the overall space complexity, it can be parallelized, thereby reducing the practical running time of density estimation.

Given the $m$-th subsample, we next apply a base partitioning algorithm to the subsampled points. 
The resulting partition can then be incorporated into EM-based density estimation in two distinct ways.
The first approach, which we refer to as \emph{global density estimation}, fits a single GMM directly to the entire subsample, where the key GMM parameters are initialized using the partition information.
This approach directly follows the EM initialization strategy adopted by
\texttt{mclust} \cite{mclust} and \texttt{scikit-learn} \cite{scikit-learn}.
The second approach, which we refer to as \emph{localized density estimation}, adopts a divide-and-conquer strategy: it fits separate GMMs to probability-weighted subsets of the subsample associated with each partition using weighted EM \cite{mclust,ohagan2019investigation}, then combines these local, partition-specific density estimates into a single estimate for the whole replicate.
The probability-weighting scheme accounts for uncertainty in point assignments near partition boundaries, allowing this uncertainty to be incorporated into each partition's local density estimate \cite{wen2026probabilistic}.
Between these two approaches, global density estimation is well established, with centroid-based algorithms and hierarchical clustering with Ward linkage expected to be particularly appropriate base partitioning methods in this setting, because their partitions are naturally aligned with Gaussian mixture assumptions \cite{bishop2006pattern,celeux1992classification,ward1963hierarchical,banfield1993model}. 
In contrast, localized density estimation is more flexible and can be applied with any base partitioning method, including graph- and density-based methods. 
Nevertheless, there is currently no consensus on a single optimal probability-weighting scheme.

To diversify the EM estimates across replicates, we can either use random starts from the same base partitioning algorithm (e.g., $k$-means) or vary the choice of base partitioning algorithm across replicates.
Using multiple initialization encourages different replicates to explore different regions of the likelihood surface. 
As a result, the ensemble as a whole becomes less sensitive to any single replicate converging to an unrepresentative local optimum.

The details on the implementation of the ensemble density estimation procedure are provided in Section 1 of the Supplementary Notes.

\subsection{Construction of Cluster Hierarchy}

Constructing a cluster hierarchy from estimated Gaussian components provides a systematic way to identify, combine, and translate density components into cluster structures across multiple resolutions. 
The core computational challenge at this stage is determining how Gaussian and mixture of Gaussian components should be merged. 
Various solutions have been proposed in the literature \cite{hennig2010merging,baudry2010combining}; however, none is fully satisfactory in terms of computational efficiency, interpretability, and applicability to large-scale datasets.

Conceptually, we find misclassification probability to be a natural way to formalize the separation between clusters composed of Gaussian components. 
In the existing literature, misclassification probability is most commonly defined as the Bayes risk of the optimal classifier under 0-1 loss for a given clustering configuration, where the optimal classifier is the deterministic decision rule that assigns each data point to the cluster with the highest posterior classification probability. 
It quantifies cluster separation by measuring the overall probability that a data point generated from any cluster would be assigned to a different cluster. 
Thus, small misclassification probability indicates well-separated clusters, whereas large misclassification probability reflects substantial overlap and poor distinguishability among some clusters.
However, this quantity is notoriously difficult to compute in practice, posing a major obstacle to scalability in our setting.
The Bhattacharyya distance can be used to derive an upper bound on this misclassification probability between two clusters. Still, it is available in closed form only for pairs of Gaussian distributions and does not extend directly to pairs of Gaussian mixtures \cite{hennig2010merging}.

Recently, we studied a new version of misclassification probability, denoted by $\pmc$, under 0-1 loss induced by an alternative randomized classifier that assigns each data point to a cluster by sampling from the corresponding posterior classification probabilities \cite{wen2026probabilistic}. 
This randomized classifier is widely used for Boltzmann exploration in sequential decision-making and reinforcement learning \cite{browne2012survey,sutton2018reinforcement}. 
Although the randomized classifier is suboptimal in the sense that it yields a misclassification probability no smaller than that of the optimal deterministic classifier for the same cluster configuration, we find that the two quantities are numerically close when measuring separation between Gaussian and mixture of Gaussian components, especially in partially overlapping and well-separated scenarios. 
Moreover, in our proposed clustering framework, the classifier is implicit and fixed, and misclassification probabilities are used only to quantify the separation of different cluster configurations, making comparison with the optimal classifier irrelevant. 

Most relevant to our problem, $\pmc$ has a particularly useful computational property: it admits a pairwise decomposition, i.e.,  
\begin{equation}\label{pmc.decomp}
\pmc = \sum_{i<j} \Delta \pmc^{(i,j)}.
\end{equation}
Here, $\Delta \pmc^{(i,j)}$ denotes the reduction in the overall $\pmc$ obtained by merging the existing clusters $i$ and $j$.
This decomposition holds for general cluster-generating distributions, including Gaussian distributions and mixtures of Gaussians (see Proposition 1 in \cite{wen2026probabilistic} for details).
Moreover, once clusters $i$ and $j$ are merged into a new cluster $i \oplus j$, the pairwise contribution between this new cluster and any existing cluster $k$ can be conveniently updated as
\begin{equation}\label{dpmc.update}
\Delta \pmc^{(i \oplus j, k)} = \Delta \pmc^{(i, k)} + \Delta \pmc^{(j, k)}.
\end{equation}
For constructing a cluster hierarchy from ensemble density estimates, it then becomes appealing to define a linkage function based on $\{\Delta \pmc^{(i,j)}\}$, which can be used to assess and rank the separation between all pairs of current clusters at each agglomerative iteration. 

In the simplest case, the linkage function
\begin{equation}
d(i,j) = -\log \Delta \pmc^{(i,j)}
\end{equation}
yields a cluster hierarchy that greedily reduces the overall $\pmc$ at each merging step.
However, in our setting, the hierarchy is constructed from an ensemble density estimate that may contain many highly similar, low-weight density components, which arise naturally when independently fitted replicates are pooled.
Because 
\begin{equation}
\Delta \pmc^{(i,j)} \propto \alpha_i \alpha_j,
\end{equation}
the raw linkage $d(i,j)$ systematically disadvantages exactly these redundant pairs: a small $\alpha_i \alpha_j$ suppresses $\Delta \pmc^{(i,j)}$ regardless of how strongly components $i$ and $j$ overlap, thereby delaying their merge.
To address this issue, we rescale $\Delta \pmc^{(i,j)}$ to remove its dependence on mixing proportions, so that near-duplicate components are merged early because they are near-duplicates, not because of their weight. 
More specifically, we define
\begin{align}
& \Delta \pmc^{*(i,j)} = \frac{\Delta \pmc^{(i,j)}}{\alpha_i \alpha_j}, \\
\intertext{and}
& d^*(i,j) = \log \left( 1 + \frac{1}{\Delta \pmc^{*(i,j)}} \right). \label{default.linkage}
\end{align}
The scaled quantity $\Delta \pmc^{*(i,j)}$ upper bounds the Bhattacharyya coefficient between the density components in clusters $i$ and $j$ \cite{wen2026probabilistic}.
We refer to the agglomerative algorithm built on linkage functions derived from $\Delta \pmc$ as the $\pmc$ hierarchical merging (PHM) algorithm.
By default, we use Eqn.~(\ref{default.linkage}) as the linkage function to construct the cluster hierarchy from the ensemble density estimate.

The constructed hierarchy provides valuable information for cluster analysis. Compared with a flat representation, it naturally reveals nested cluster structure and allows users to compare grouping patterns at multiple resolutions. Visualization tools, including heatmaps and dendrograms, further help users explore these multiresolution patterns. Finally, although we do not advocate selecting a single flat clustering as the default representation, one can be obtained, when desired, by applying a threshold to $\pmc$.

\subsection{Computational Complexity}

Ensemble density estimation scales linearly with the subsample size $\rho N$.
Because the $M$ replicates are built from independent subsamples and initializations, they are embarrassingly parallel \cite{moler1986matrix}. 
Once model fitting by EM is complete, only the parameters of the $K$ pooled atomic components need to be retained; neither the subsamples nor their responsibility matrices are required. 
Consequently, the persistent memory requirement for ensemble density estimation is independent of $N$.

For cluster hierarchy construction, evaluating pairwise $\Delta \pmc$ values and performing PHM merging have time complexity $O(SKd^2 + SK^2)$ and $O(K^2)$, respectively, where $S$ is a fixed Monte Carlo sampling budget.
When $K$ is large, we apply a preprocessing procedure that identifies and consolidates highly overlapping Gaussian components, thereby reducing the number of atomic components $K$ and further lowering the computational cost.
The persistent space complexity of the PHM algorithm is $O(K^2)$, which is completely decoupled from $N$.
In comparison, prior merging approaches \cite{baudry2010combining,hennig2010merging} must keep an $N$-sized responsibility matrix resident throughout the merge process, resulting in $O(NK^2)$ time and $O(NK)$ space complexity.
Because $N \gg K$ in large-scale datasets, such as atlas-scale single-cell RNA-seq data, the proposed PHM algorithm becomes a substantially more efficient and scalable alternative than existing methods.

\section{Results}

\subsection{Synthetic Data Examples}

We demonstrate PHM-ensemble on a series of synthetic datasets to evaluate its fidelity, interpretability, stability, and scalability and to benchmark its performance against existing clustering methods.

\begin{figure} [H]
    \centering
    \begin{subfigure}[b]{0.55\textwidth}
      \centering
      \includegraphics[width=\textwidth]{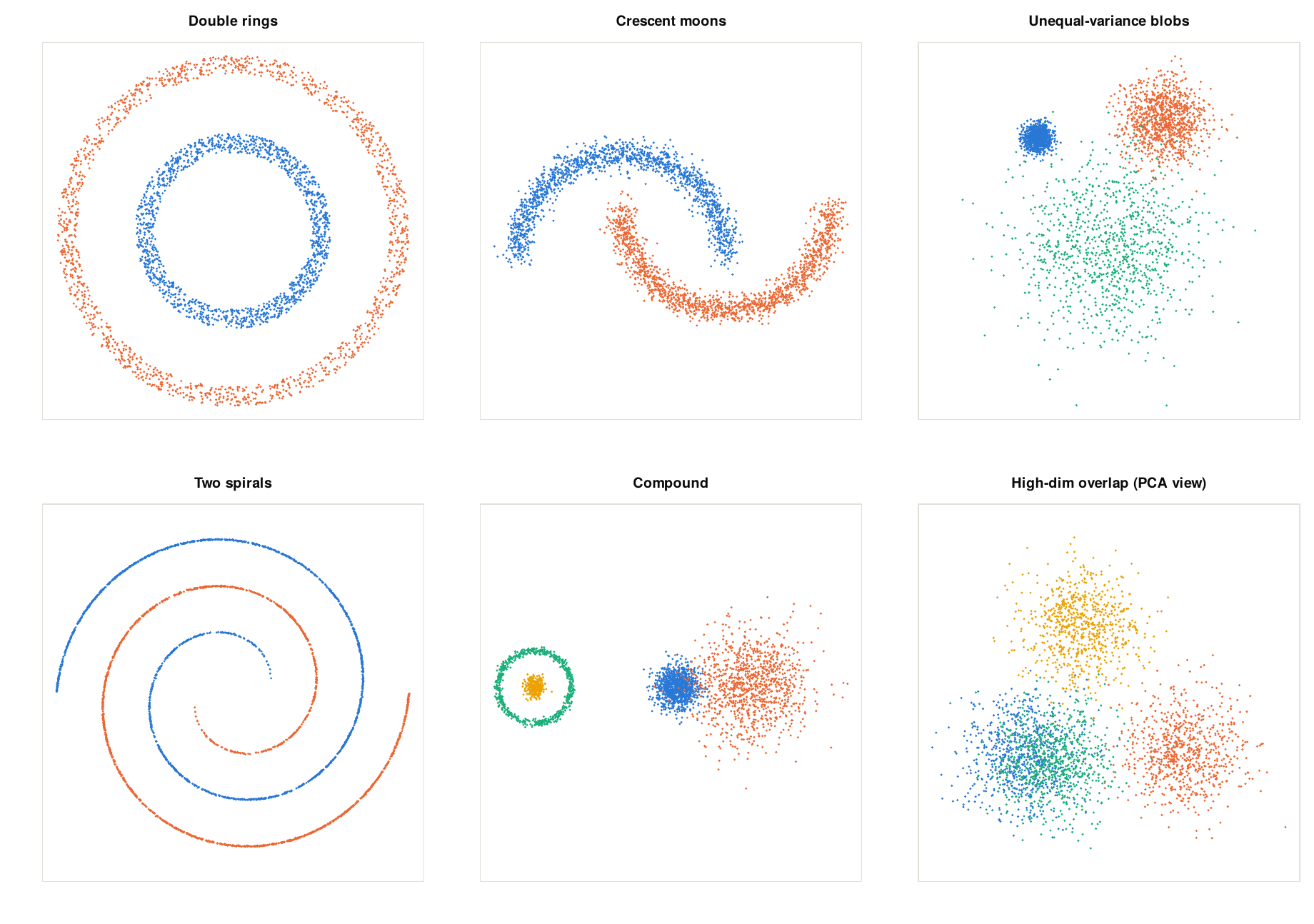}
      \caption{~}
    \label{sim_figure_a}
    \end{subfigure}
\begin{subfigure}[b]{0.38\textwidth}
    \centering
    \includegraphics[width=\textwidth]{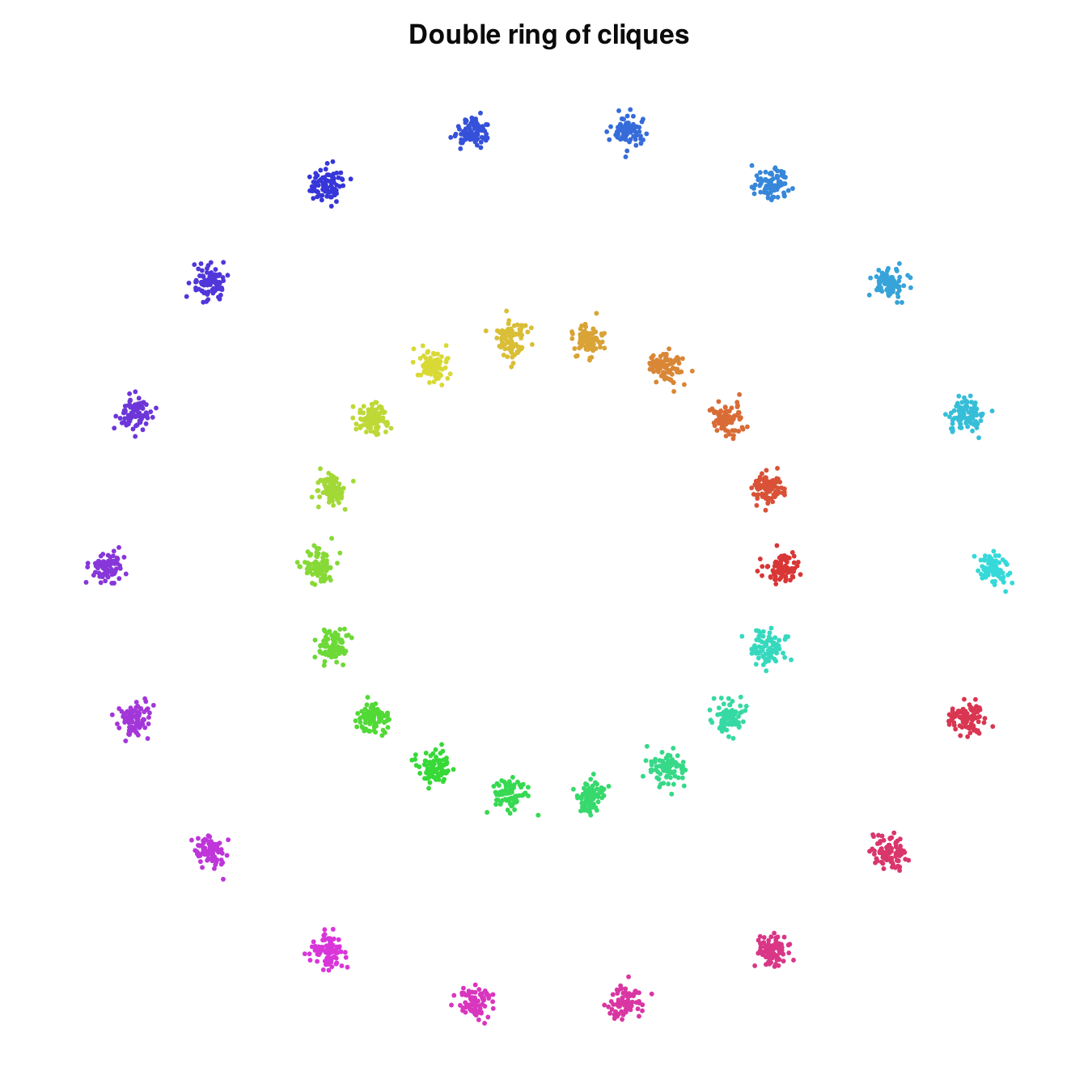}
    \caption{~}
    \label{sim_figure_b}
\end{subfigure}
 \caption{\textbf{Synthetic datasets for evaluating clustering performance}~ (A) Six synthetic datasets designed to challenge existing clustering methods. (B) A synthetic dataset with a clear multiresolution structure, consisting of two well-separated rings at the coarser resolution, with each ring further divided into 18 distinct cliques of points at the finer resolution.}

\end{figure}

The first set of synthetic datasets consists of six challenging clustering problems (Figure \ref{sim_figure_a}) designed to highlight limitations of various existing clustering methods (Section 2.1 of the Supplementary Notes). 
The first five datasets contain two-dimensional points and feature clusters that are either non-convex (i.e., double rings, crescent moons, and two spirals), exhibit different levels of within-cluster heterogeneity (i.e., unequal-variance blobs), or combine both characteristics (i.e., the ``compound'').
The sixth dataset consists of points in a 60-dimensional space that form overlapping clusters with unequal variances and covariance structures.
Each dataset contains 3,000 points with a relatively high signal-to-noise ratio, as the cluster patterns are clearly visible to the human eye. 
We apply the PHM-ensemble procedure with $M=5$ and $\rho=1.0$ to all six datasets. 
In each ensemble replicate, we perform global density estimation, allowing up to 100 Gaussian components and initializing the EM algorithm using the randomized hierarchical clustering procedure implemented in \texttt{mclust}.
For comparison, we also apply seven commonly used clustering methods: $k$-means, hierarchical clustering with Ward and single linkage, model-based clustering using \texttt{mclust}, density-based clustering using HDBSCAN and OPTICS, and community detection using the Leiden method. 
To evaluate clustering fidelity, we fix the number of clusters at the true number for each dataset and compute the adjusted Rand index (ARI) between the estimated cluster assignments and the known ground-truth labels for each method.
The results are summarized in Table \ref{sim_ari_tab}. 
PHM-ensemble and Leiden are the only two methods that maintain consistently high ARIs across all six datasets, demonstrating their ability to faithfully recover the underlying cluster structures across a range of challenging scenarios. 
Additionally, the performance of PHM-ensemble remains virtually unchanged when the subsampling proportion $\rho$ is reduced to 0.75, 0.50, and 0.25 while keeping $M=5$ (Supplementary Table 1).
In contrast, each of the remaining methods performs poorly on at least one dataset, highlighting their sensitivity to particular types of cluster structure.
For example, \texttt{mclust}, $k$-means, and hierarchical clustering with Ward linkage all fail to recover the non-convex cluster structures in the double rings, crescent moons, two spirals, and compound datasets. 
The density-based methods HDBSCAN and OPTICS, on the other hand, struggle even with a modest increase in dimensionality: although all methods are applied to the ``high-dim overlap'' dataset using only the first four principal components (PCs) of the 60-dimensional data (selected by a scree test), their performance deteriorates substantially due to the loss of local density contrast that occurs as dimensionality grows, which is a well-known fundamental challenge for density-based clustering methods.

\begin{table}[!ht]
\centering
\caption{Adjusted Rand Index (ARI) of eight clustering methods on the six synthetic datasets.}
\label{sim_ari_tab}
\resizebox{\textwidth}{!}{%
\begin{tabular}{lcccccc}
\toprule
Method & Double Rings & Crescent Moons & Unequal-Variance Blobs & Two Spirals & Compound & High-Dim Overlap \\
\midrule
k-means                        & 0.001          & 0.257          & 0.844          & 0.029          & 0.662          & 0.955          \\
Hierarchical (Ward)                & 0.000          & 0.191          & 0.942          & 0.037          & 0.693          & 0.909          \\
Hierarchical (single)              & 1.000 & 1.000 & 0.000          & 1.000 & 0.474          & 0.000          \\
mclust                    & 0.000          & 0.503          & 0.961 & 0.046          & 0.776          & 0.956          \\
OPTICS                             & 0.000          & 0.995          & 0.928          & 0.977          & 0.500          & 0.006          \\
HDBSCAN                            & 1.000 & 0.965          & 0.807          & 0.964          & 0.785          & 0.230          \\
Leiden                             & 1.000 & 1.000 & 0.950          & 1.000 & 0.940 & 0.954          \\
PHM-ensemble                       & 1.000 & 1.000          & 0.961          & 1.000 & 0.938          & 0.957 \\
\bottomrule
\end{tabular}%
}
\end{table}

\begin{figure}[H]
    \centering
    \includegraphics[width=0.65\textwidth]{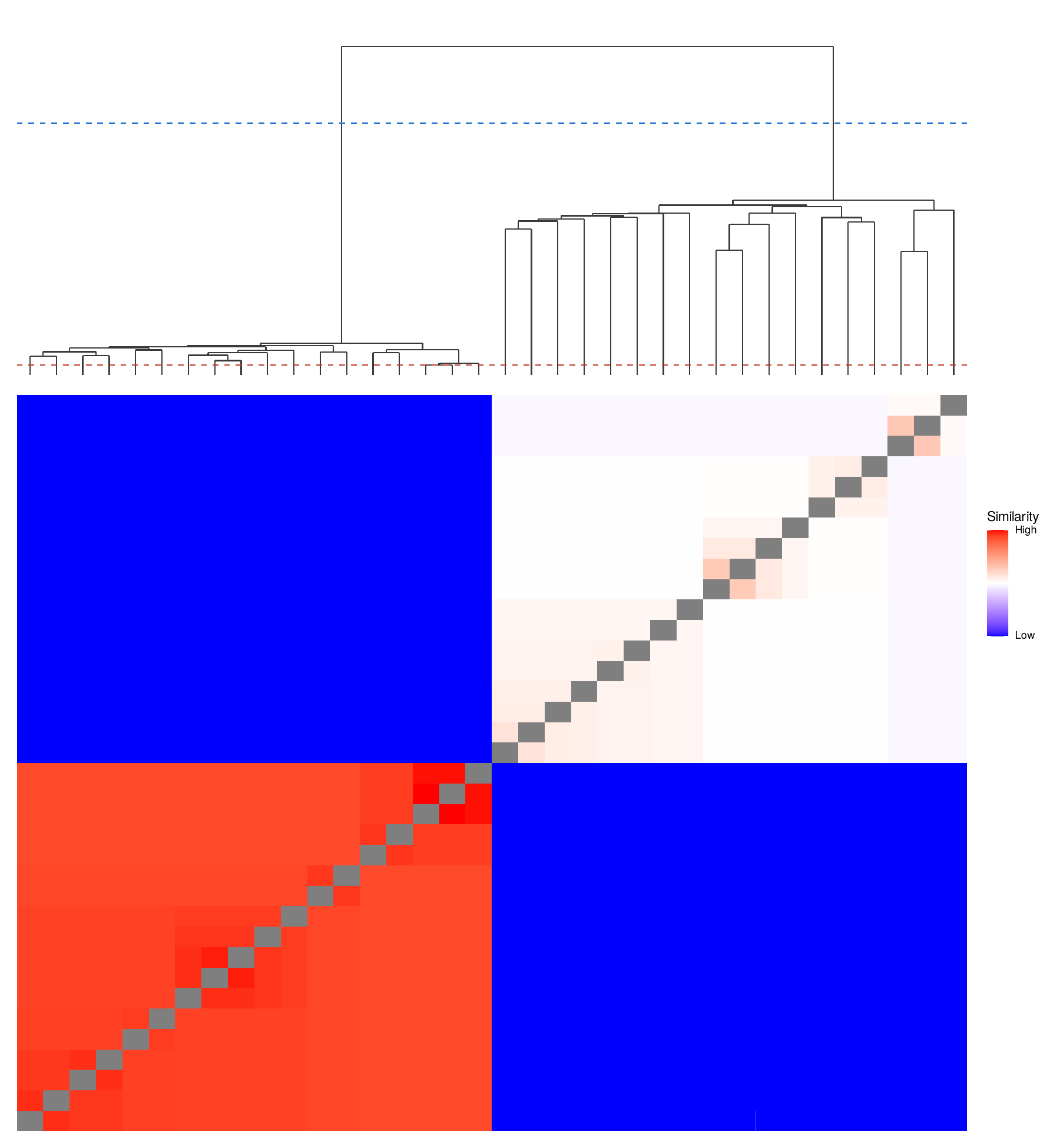}
    \caption{
        \textbf{PHM-ensemble recovers multiresolution structure in the double rings of cliques dataset}~ The PHM-ensemble result, obtained with $M=5$ and $\rho=1.0$, correctly recovers both the two-ring structure at the coarser resolution (blue dotted line) and the 36 individual cliques at the finer resolution (red dotted line). The resulting dendrogram and heatmap provide a direct and interpretable representation of the multiresolution structure. The cliques in the inner ring are closer to each other than those in the outer ring, which is reflected in the dendrogram by the shorter branch lengths in the inner ring.}
    \label{phm_double_ring_clique}
\end{figure}

Next, we evaluate interpretability using another simulated dataset, termed ``double rings of cliques'', consisting of 3,000 two-dimensional data points with a clear multiresolution structure (Figure \ref{sim_figure_b}).  
At the coarser resolution, the dataset consists of two well-separated rings, whereas at the finer resolution, each ring is further divided into 18 distinct cliques of points. 
The PHM-ensemble result, obtained with $M=5$ and $\rho=1.0$, correctly recovers both levels of structure, which is directly represented and readily interpretable from the resulting dendrogram and heatmap (Figure \ref{phm_double_ring_clique}).
Reducing the subsampling proportion $\rho$ to 0.5 or 0.1 yields virtually identical results.
In contrast, Leiden recovers the 36 individual cliques across nearly the entire range of the resolution parameter but fails to recognize the two-ring structure at any resolution. 
This is likely because the constructed graph contains no edges connecting distinct cliques, preventing modularity-based community detection methods from merging them regardless of the choice of resolution parameter.

\begin{figure}[H]
    \centering
    \begin{subfigure}[c]{0.85\textwidth}
    \includegraphics[width=\textwidth]{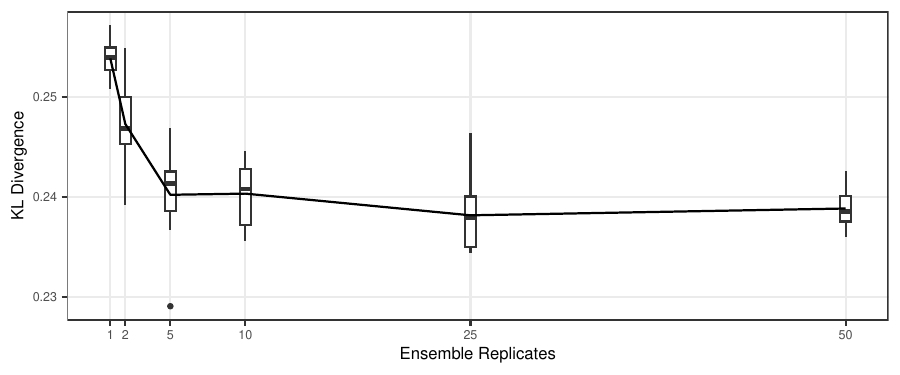}
    \caption{Double Rings} \label{stability_fig_a}
    \end{subfigure}
    
    \vskip 7pt

     \begin{subfigure}[c]{0.85\textwidth}
    \includegraphics[width=\textwidth]{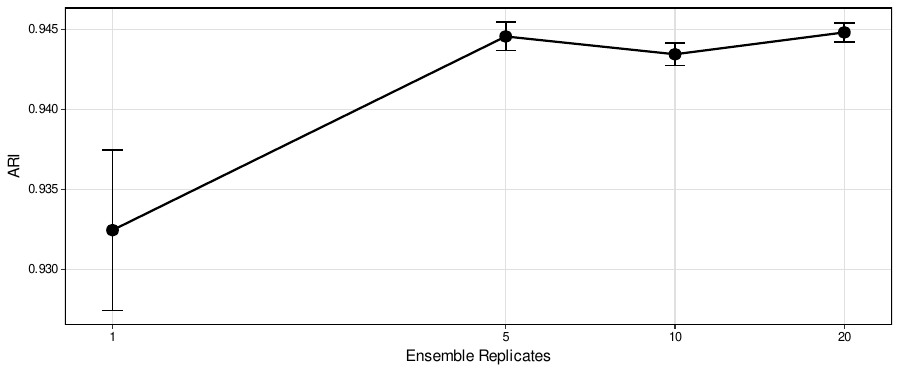}
    \caption{Compound} \label{stability_fig_b}
    \end{subfigure}
    
    \caption{ \textbf{Ensemble improves density estimation and clustering stability}~ (A) Boxplots for the KL divergence of the global density estimation procedure for the double ring data as the ensemble size ($M$) increases. The solid line traces the mean KL divergence. We observe a consistent decrease in KL divergence as $M$ increases, suggesting that the ensemble density estimation more accurately captures the underlying density.~~(B) ARI of PHM-ensemble on the compound data as the ensemble size ($M$) increases. The solid line traces the mean ARI, and vertical bars indicate $\pm 1$ standard deviation across runs.}
    \label{rings_kl_fig}
\end{figure}

To assess the stability of the PHM-ensemble and the effect of its ensemble density estimation procedure, we first focus on the double-ring data from the first set of synthetic datasets, where the simulated data points are uniformly distributed over two annuli. 
We compare the KL divergence between the uniform distribution and the density estimates by varying the number of replicates ($M = 1, 2, 5, 10, 25, 50$) using both global and localized density estimation approaches. 
For global density estimation, each replicate uses 24 Gaussian components. 
For localized density estimation, we initialize each replicate using $k$-means with $K=6$ partitions and allow up to four Gaussian components for estimating each partition-specific density. 
We repeat the experiment 10 times under each estimation setting to quantify the variability in the resulting KL divergence.
For both density estimation procedures, we observe that estimation accuracy improves as $M$ increases, while the variability of the estimates decreases (Figure \ref{stability_fig_a}, Supplementary Figure 1). 
This behavior is consistent with our design expectation for ensemble density estimation, as combining multiple replicates enables the model to capture the true data density more accurately and reliably, even when the model fitted in each replicate is misspecified.
To further demonstrate that improvements in ensemble density estimation can translate into better clustering performance, we turn to the ``compound'' dataset, for which none of the evaluated methods perfectly recovers the ground-truth clustering.
We compare the ARIs obtained by PHM-ensemble with $M=1, 5, 10$, and $20$, while fixing $\rho=1$ (Figure \ref{stability_fig_b}).
As $M$ increases, both clustering accuracy and stability improve, as reflected by higher ARIs and reduced variability across repeated runs.

\begin{figure}[H]
    \centering
    \includegraphics[width=0.75\textwidth]{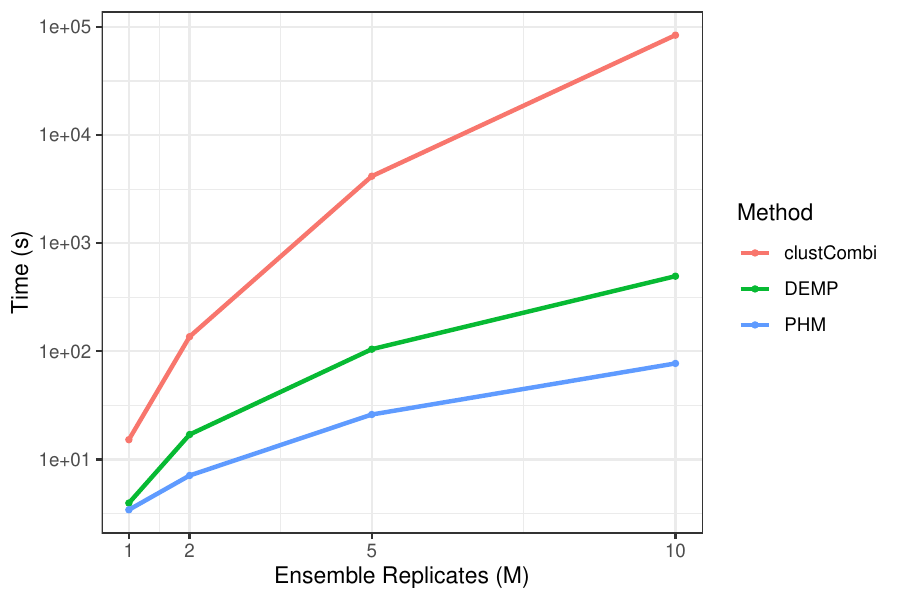}
    \caption{
        \textbf{Runtime for constructing cluster hierarchy using PHM, DEMP, and clustCombi}~ Wall-clock times for the PHM, DEMP, and \texttt{clustCombi} procedures applied to the global ensemble density estimates are shown. Each plotted point represents the average wall-clock time over 10 repeated runs under the same setting. The ensemble sizes $M=1, 2, 5,$ and $10$ correspond to mixture densities containing $24, 48, 120,$ and $240$ components, respectively. Among the three methods, PHM exhibits the best scalability as the ensemble size increases.
    }
    \label{rings_timing_fig}
\end{figure}

Finally, to evaluate scalability, we report the wall-clock time required to construct the clustering hierarchy in the double ring data using PHM and two alternative procedures, \texttt{clustCombi} \cite{baudry2010combining}, and DEMP/DEMP+ \cite{hennig2010merging, melnykov2016merging} (Figure \ref{rings_timing_fig}).
All three methods can merge the Gaussian components from the ensemble density estimates to construct a clustering hierarchy and recover the double-ring structure at the resolution of $K=2$ across density estimates obtained with varying numbers of ensemble replicates (Supplementary Figures 2 and 3).
Both PHM and DEMP identify $K=2$ as the most prominent structural change during the merging process, whereas the built-in selection procedure of \texttt{clustCombi} favors solutions with a larger number of clusters (Supplementary Notes 2.2).
Among the three methods, PHM exhibits the best scalability as the number of ensemble replicates increases, requiring less than 100 seconds for $M=10$, corresponding to 240 mixture components.
In contrast, \texttt{clustCombi} becomes computationally expensive even with a moderate number of ensemble replicates, requiring more than an hour to complete when $M=10$.
DEMP scales substantially better than \texttt{clustCombi} but remains less scalable than PHM as the number of mixture components increases: at $M=10$, DEMP takes 8.3 minutes, compared to PHM's 1.3 minutes.

\subsection{MNIST Handwritten Digits}

In this illustration, we apply PHM-ensemble to the MNIST handwritten digit dataset \cite{lecun1998gradient}.
The dataset contains 70,000 grayscale images of handwritten digits from 0 to 9, each represented by a $28 \times 28$ pixel grid, and is widely used as a benchmark for clustering methods.
We flatten each image into a vector, retain the top 100 principal components (PCs), and embed the resulting data in two dimensions using t-SNE, following the common practice of PCA preprocessing before t-SNE \cite{vandermaaten2008visualizing,vandermaaten2014accelerating}.
The resulting t-SNE embedding is then used as the input for clustering (Section 3 of the Supplementary Notes).

We evaluate PHM-ensemble with both global and localized density estimation, using $M=10$ replicates per ensemble and a subsampling proportion of $\rho=0.1$ per replicate. 
To benchmark its performance, we also evaluate $k$-means, hierarchical clustering (Ward linkage), Leiden clustering, \texttt{mclust}, HDBSCAN, and SC3s (an ensemble consensus clustering method based on $k$-means)\cite{kiselev2017sc3}. 
We assess the quality of the clustering results against the ground-truth digit labels using clustering accuracy (ACC), normalized mutual information (NMI), and ARI.
We compute these metrics for a 10-cluster solution from each method, except HDBSCAN, for which we use an 11-cluster solution.
Specifically, we set $K=10$ for $k$-means, \texttt{mclust}, and SC3s; tune the Leiden resolution parameter to obtain exactly 10 communities; and cut the dendrograms produced by hierarchical clustering and PHM-ensemble at heights that yield 10 clusters. For HDBSCAN, no value of the minimum cluster size parameter yields exactly 10 clusters, so we use the smallest integer value that yields 11 clusters.
All methods except hierarchical clustering and HDBSCAN involve stochastic components, including random initialization in $k$-means, \texttt{mclust}, and SC3s, subsampling in PHM-ensemble, and the stochastic local-moving heuristic in Leiden clustering. 
To assess their stability, we run each randomized algorithm 100 times with different random seeds and examine the variability in the resulting performance metrics. 
Additional details on the clustering procedures are provided in Section 3 of the Supplementary Notes.

\begin{figure}[H]
    \centering
    \includegraphics[width=0.9\textwidth]{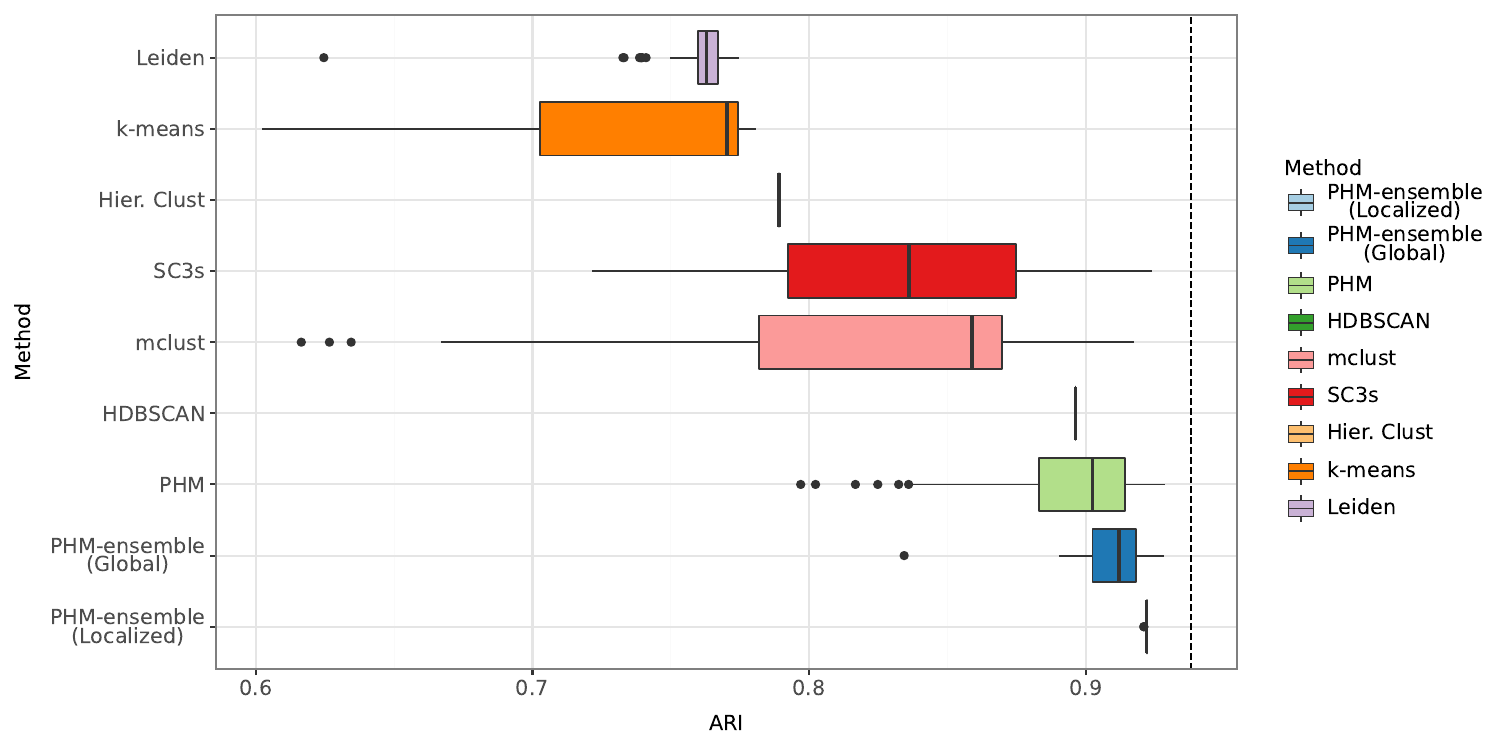}
    \caption{
        \textbf{PHM-ensemble achieves high clustering fidelity and stability on MNIST data}~ Boxplots of ARI values for the methods applied to the MNIST dataset. The oracle classifier performance is shown by the dashed vertical line. The PHM methods outperform the other baseline methods, with PHM-ensemble having very consistent performance across independent runs.
    }
    \label{mnist_compare_fig}
\end{figure}

Relative to the ground-truth labels, PHM-ensemble achieves the best performance across all three metrics among the evaluated clustering methods (Figure \ref{mnist_compare_fig} and Supplementary Table 2). 
Its accuracy approaches that of an oracle classifier, which estimates each digit class's density separately using a Gaussian mixture model fitted with the ground-truth labels and assigns each image to the digit class with the highest posterior probability.
Global and localized density estimation yield similar results, suggesting that both ensemble density estimates accurately capture the underlying data distribution.

PHM-ensemble also yields the most stable results among the randomized clustering algorithms (Figure \ref{mnist_compare_fig} and Supplementary Table 2). 
To assess whether this improvement in stability is attributable to the ensemble strategy, we also evaluate the original PHM procedure, which uses a single density estimate fitted to the full dataset and is equivalent to PHM-ensemble with $M=1$ and $\rho=1$. 
Compared with the original PHM procedure, PHM-ensemble improves both accuracy (mean ARI increases from 0.891 to 0.910) and stability (the standard deviation of ARI decreases from 0.030 to 0.013).
Moreover, it is worth noting that the reduction in algorithmic variability achieved by PHM-ensemble is considerably greater than that achieved by SC3s, which also employs an ensemble strategy but aggregates partitions rather than density estimates.

\begin{figure}[H]
    \centering
    \includegraphics[width=0.95\textwidth]{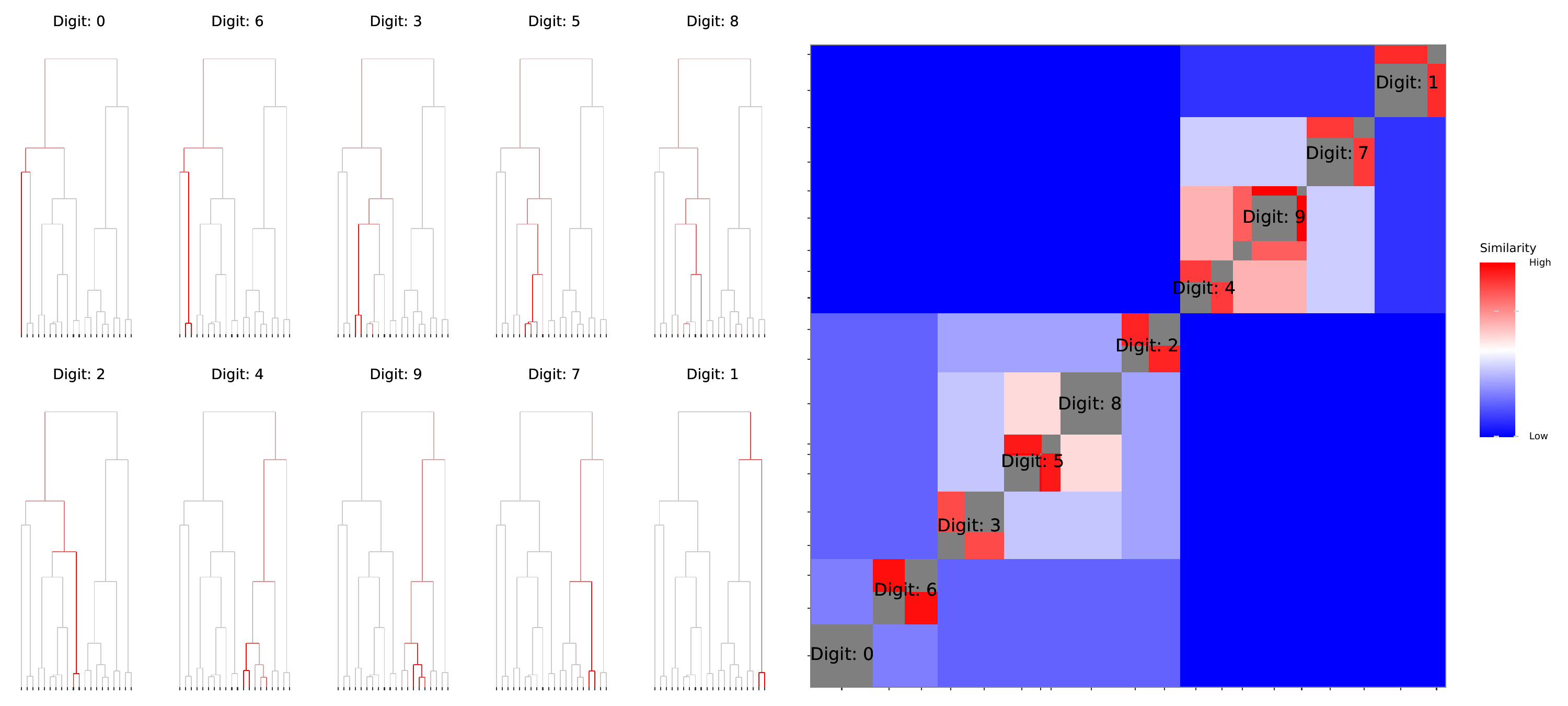}
    \caption{
        \textbf{PHM uncovers multiresolution structure in MNIST}~ Dendrograms for the PHM merging procedure from the analysis of MNIST data, showing the top 20 clusters. The colors of the dendrograms correspond to the proportions of observations in a given branch belonging to a given digit class. Both the dendrogram heights and the heatmap colors recover components corresponding to unique digits and show relationships between digits.
    }
    \label{mnist_phm_fig}
\end{figure}

Finally, PHM-ensemble can reveal complex multiresolution structure in the data beyond a single flat partition, providing enhanced interpretability compared with existing methods. Figure \ref{mnist_phm_fig} presents the PHM dendrogram and heatmap from one PHM-ensemble run, displaying the top 20 clusters in the hierarchy. The visualization shows that the 10 clusters corresponding to the individual digits are generally well separated, while also revealing substantial heterogeneity within several digit clusters, with subclusters corresponding to distinct writing styles.
For example, the cluster representing the digit 2 contains two distinct subclusters: one corresponding to the conventional form, $2$, and the other to a form written with a loop at the bottom, {\cursive 2}. Subclusters within other digits, including 1, 4, 7, and 9, similarly capture variation in writing style, such as upright versus slanted forms, for example, $9$ versus \textit{9} (Supplementary Figure 4).
At coarser resolutions, the hierarchy further reveals that the 10 digit clusters are not equally separated, with certain groups of handwritten digits exhibiting greater structural similarity. At the highest level of the hierarchy, the dendrogram separates the digits into two broad groups: one consisting primarily of the straight-line digits 1, 4, 7, and 9, and the other consisting of the more curved digits 0, 2, 3, 5, 6, and 8. The heatmap further indicates relatively strong similarities among 9, 4, and 7, as well as among 8, 5, and 3. The shared structural characteristics of 9 and 4 have previously been noted in \cite{goggin2024eschr}. Here, PHM-ensemble reveals these and other relationships among handwritten digits in a comprehensive and systematic manner.

\subsection{Human Immune Health Atlas Single-cell RNA-seq Data}

In this illustration, we apply the PHM-ensemble to an atlas-scale single-cell RNA-sequencing (scRNA-seq) dataset from the Human Immune Health Atlas \cite{gong2024longitudinal} to perform cluster analysis. The atlas collects 2.09 million peripheral blood mononuclear cells, with approximately 33,500 genes measured in each cell. 
We download the quality-controlled, dimension-reduced data directly from the original publication. 
The final dataset used for cluster analysis contains 1.82 million cells (after QC and removing doublet and contaminant cells), each represented by 30 principal components (PCs). 
Using these data, we fully reproduce the Leiden clustering analysis reported in \cite{gong2024longitudinal}.

We apply the PHM-ensemble procedure with global density estimation to perform cluster analysis of this dataset. 
We use $M=50$ ensemble replicates, each comprising 100,000 subsampled observations (or 5.5\% of the complete dataset). 
For the density estimation of each replicate, we allow up to 100 mixture components. 
The resulting ensemble mixture contains 2,706 components in total.

\begin{figure}[H]
    \centering
   
    \includegraphics[width=1\textwidth]{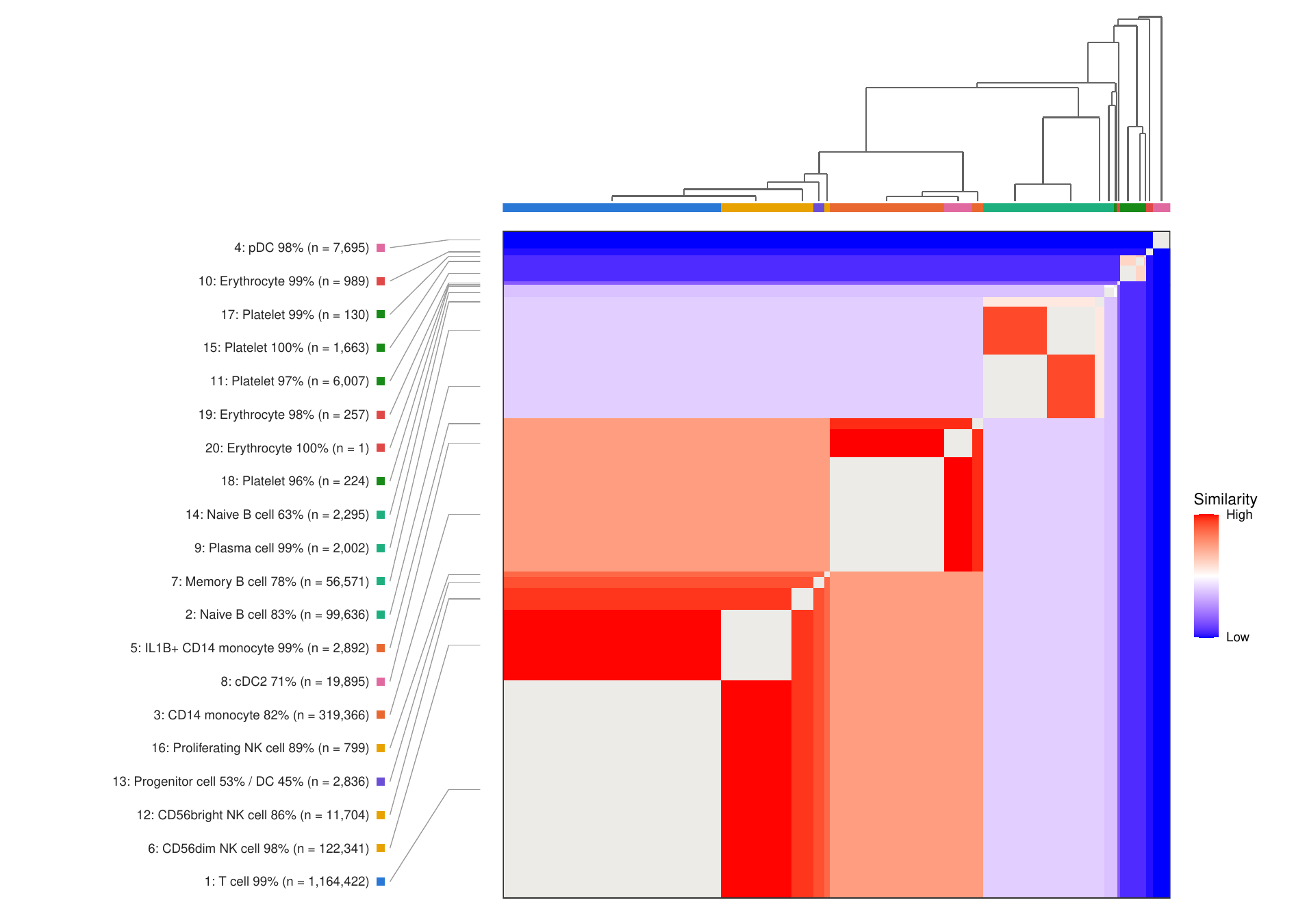} \\
   
    \caption{
        \textbf{PHM recovers broad cell-type groupings and their relationships in expression space}~ Overlaid PHM dendrogram and heatmap from the PHM-ensemble analysis of the Human Immune Health Atlas at the resolution $K=20$. 
        Individual cells are assigned to clusters using the maximum a posteriori (MAP) rule. 
        Each cluster is labeled with its index, its most frequent AIFI level 1 annotation and the percentage of cells in the cluster carrying that annotation, and the total number of cells assigned to the cluster ($n$).
        Cluster indices serve only as identifiers. The area of each diagonal block is proportional to the number of cells assigned to the corresponding cluster.
        The dendrogram and heatmap reveal major cell type groups and their relationships.
    }
    \label{hiha_phm_all_fig}
\end{figure}

The PHM-ensemble result (Figure \ref{hiha_phm_all_fig}) successfully identifies clusters corresponding to major immune cell types.
For reference, we use the cell type annotations from the Allen Institute for Immunology (AIFI), which were manually curated into a three-level hierarchy through iterative Leiden subclustering and examination of canonical RNA and surface protein (CITE-seq) marker expression \cite{gong2024longitudinal}.
The PHM-ensemble result at $K=20$ achieves an ARI of 0.965 and an NMI of 0.908 relative to the level 1 cell type annotations that define 9 broad cell types (Supplementary Table 3 and Supplementary Figure 5).
In comparison, Leiden clustering achieves a maximum ARI of 0.789 (NMI $= 0.841$) at a resolution of 0.01, yielding 10 clusters, across the examined resolution range of 0.005 to 1.0 (corresponding to 9 to 36 clusters).
More importantly, the PHM results reveal complex relationships among immune cell types in gene expression space, including their internal heterogeneity and hierarchical structure. 
These relationships can be directly interpreted from the resulting dendrogram and heatmap (Figure \ref{hiha_phm_all_fig}). 

At $K=20$, we observe that clusters representing the major lineages (T/NK cells, B cells, and myeloid cells) are clearly distinct from one another: overlap between any lymphoid and myeloid cluster, as well as between T/NK and B cell clusters, is negligible.

Nearly all T cells (more than 99.99\%) fall into a single cluster (cluster 1), which forms a branch with the NK cell clusters. 
In particular, the T cell cluster overlaps most with the CD56$^{\mathrm{dim}}$ NK cluster (cluster 6), consistent with the shared cytotoxic program of NK cells and effector-memory CD8 T cells, whereas the CD56$^{\mathrm{bright}}$ (cluster 12) and proliferating NK (cluster 16) clusters join this branch later.

 Three clusters (clusters 2, 7 and 9), containing 98.5\% of all annotated B cells (and no other cell types), form a distinct branch that is well separated from all other cell types.
The two largest clusters correspond to naive B cells (cluster 2, including transitional B cells) and memory B cells (cluster 7, including effector B cells), respectively, and exhibit the only noticeable overlap within the branch.
This pattern is consistent with naive B cell differentiation into memory B cells following antigen exposure \cite{cyster2019bcell}.
Plasma cells form a distinct cluster (cluster 9) that joins this branch last, with negligible overlap with the naive and memory B cell clusters.
A fourth, small cluster (cluster 14, containing 1.4\% of B cells) consists of a mixture of annotated naive and memory B cells. This cluster is isolated from the other B cell clusters and joins the hierarchy outside the main B cell branch, alongside a small platelet cluster (cluster 18).

Monocytes and type 2 conventional dendritic cells (cDC2) occupy three adjacent clusters (clusters 3, 5 and 8) that form a myeloid block clearly separated from the lymphoid clusters.
Together, these three clusters contain 99.97\% of all annotated monocytes.
Classical CD14, nonclassical CD16, and intermediate monocytes make up the largest cluster (cluster 3) and are not separated at this resolution.
This cluster overlaps with the cDC2 cluster (cluster 8), suggesting a continuum of expression profiles between monocytes and cDC2.
The overlap is consistent with the intermediate expression profile of monocyte-like CD14$^{+}$ cDC2 (DC3) and the known difficulty of separating DC3 from CD14$^{+}$ monocytes \cite{villani2017single, bourdely2020transcriptional}.
A small cluster of IL1B$^{+}$ CD14 monocytes (cluster 5), likely reflecting an inflammatory state, is the last to join the myeloid block.
Outside this block, plasmacytoid dendritic cells (pDCs) form an isolated cluster (cluster 4), while type 1 conventional dendritic cells (cDC1) cluster with progenitor cells (cluster 13).

Other annotated level 1 cell types are rare in the clustering dataset.
Specifically, platelets, progenitor cells, erythrocytes, and innate lymphoid cells (ILCs) account for 0.43\%, 0.08\%, 0.08\%, and 0.05\% of all cells, respectively.
Platelets and erythrocytes each form a main cluster (clusters 11 and 10, respectively) composed predominantly ($\geq 96\%$) of their respective cell types.
Progenitor cells account for the majority (53\%) of a cluster (cluster 13) that also contains dendritic cells (cDC1 and ASDC).
In contrast, ILCs do not form a distinct cluster at this resolution but are instead distributed across the T cell and NK cell clusters (mainly clusters 1 and 12).

Moreover, the PHM-ensemble results enable a more detailed examination of relationships among cell types at finer resolutions by simply zooming into individual branches of the hierarchy, without the need for iterative subclustering.

\begin{figure} [H]
    \centering
    \begin{subfigure}[b]{0.47\textwidth}
      \centering
      \includegraphics[width=\textwidth]{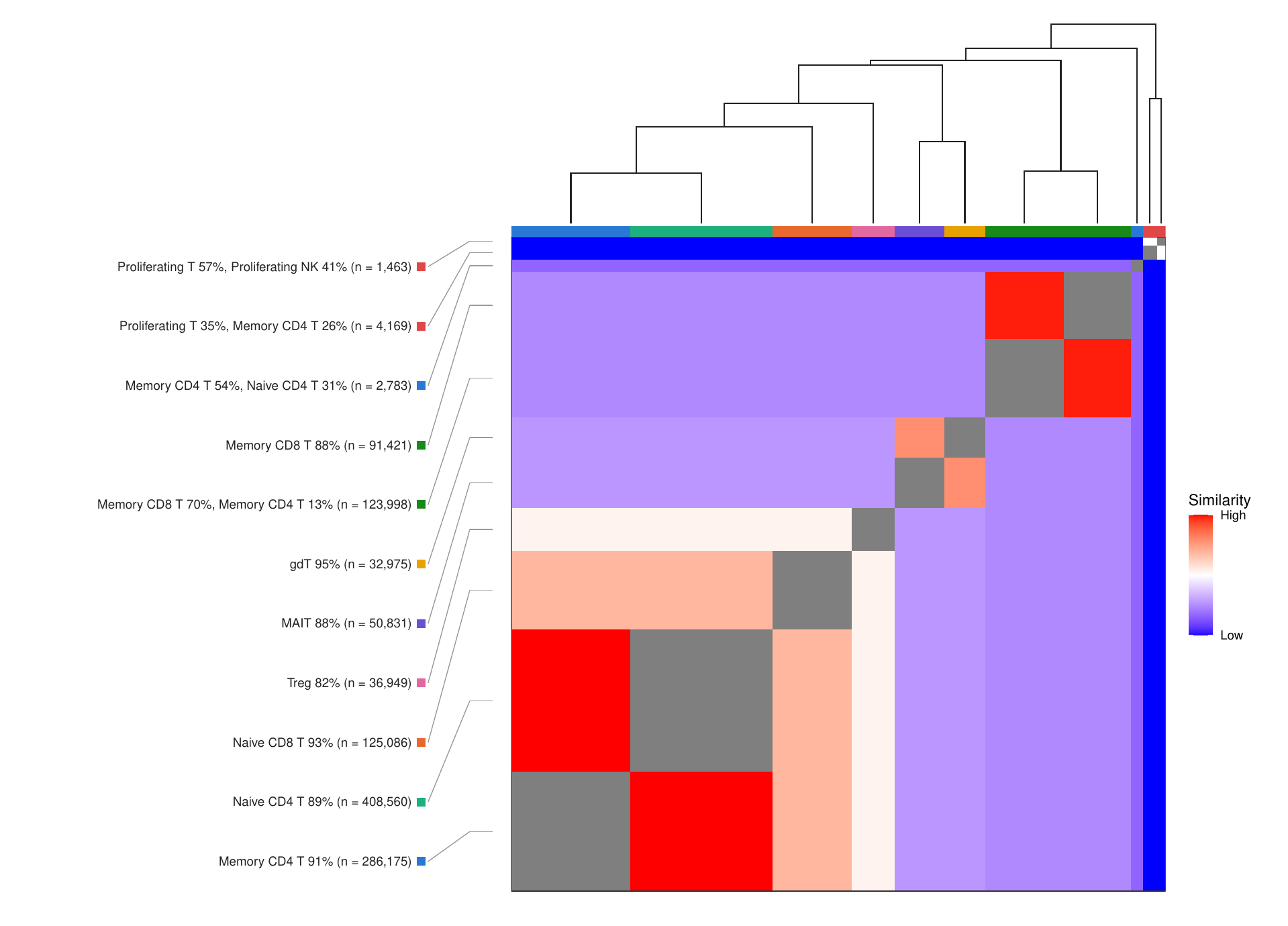}
      \caption{T cells}
    \label{hiha_t_closeup_fig}
    \end{subfigure}
\begin{subfigure}[b]{0.50\textwidth}
    \centering
    \includegraphics[width=\textwidth]{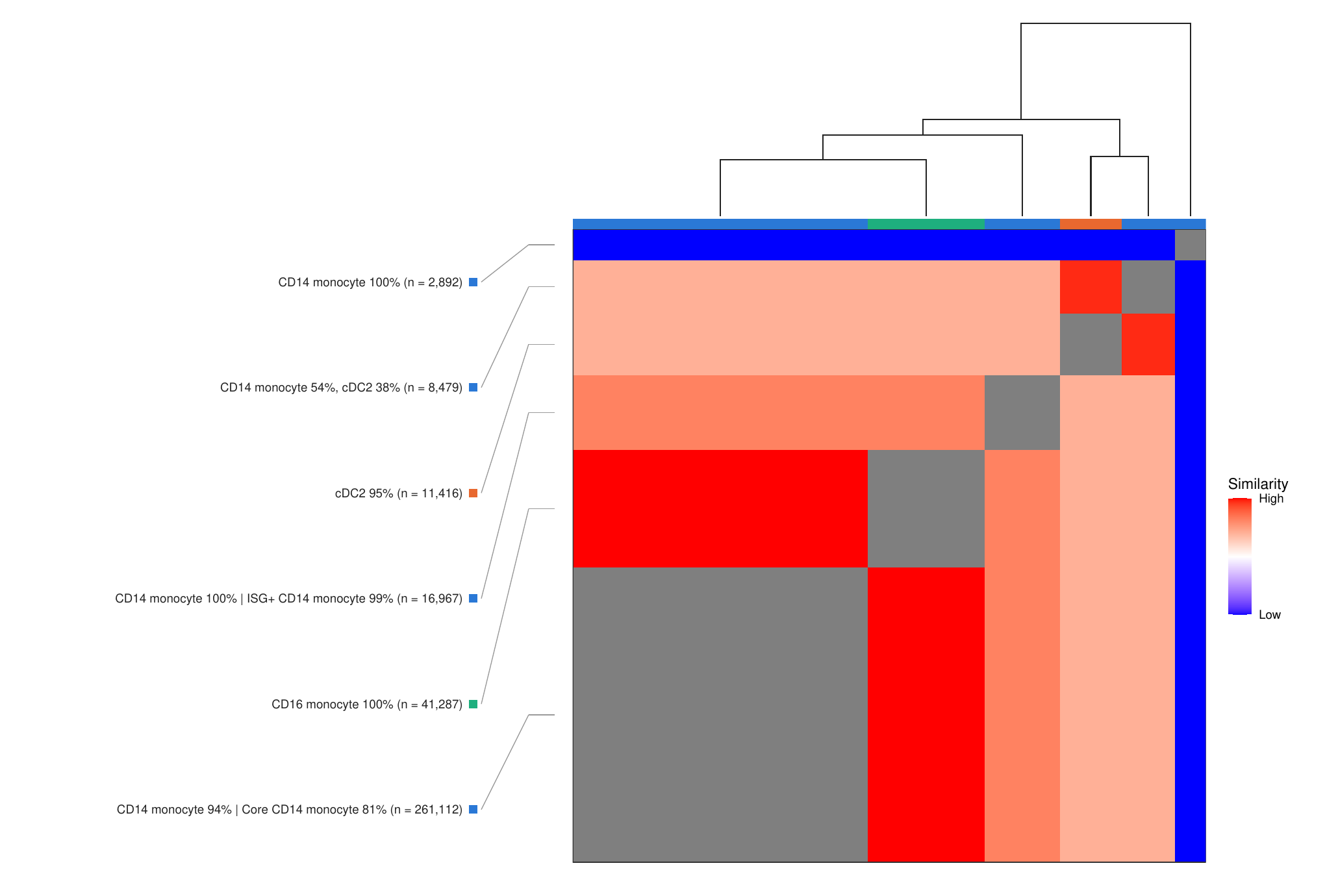}
    \caption{Monocytes}
    \label{hiha_monocyte_closeup_fig}
\end{subfigure}
 \caption{\textbf{PHM-ensemble recovers subcluster structure within T cells and Monocytes at finer resolutions}~ PHM-ensemble heatmaps and dendrograms of (A) the T cell block (cluster 1 at $K=20$), shown at $K=41$, and (B) the myeloid block (clusters 3, 5 and 8 at $K=20$), shown at $K=27$. 
 Each cluster is labeled with its most frequent AIFI level 2 annotation, the percentage of cells in the cluster carrying that annotation, and the total number of cells assigned to the cluster ($n$). A second annotation is listed when it accounts for at least 10\% of the cluster's cells. In (B), the two major CD14 monocyte clusters also show their most frequent AIFI level 3 annotation after the vertical bar.  The area of each diagonal block is proportional to the number of cells assigned to the cluster. The subgroup structures within the T cell and monocyte blocks revealed from the PHM-ensemble analysis is consistent with known biology.
 }

\end{figure}

Figure \ref{hiha_t_closeup_fig} shows a closer examination of the subtree spanning the T cell cluster (cluster 1) at $K=41$. Within the T cell cluster, the most prominent separation distinguishes a small group of proliferating cells (including some proliferating NK cells) from the remaining T cells. 
This separation appears to reflect a shared cell-cycle program rather than differences in T cell lineage.
The remaining lineage-defined T cell subclusters clearly separate into three major groups: conventional naive and CD4 T cells (naive CD4, memory CD4, naive CD8 and regulatory T cells, or Treg), innate-like T cells (MAIT and $\gamma\delta$T, or gdT), and memory CD8 T cells, including the cytotoxic effector cells.

Figure \ref{hiha_monocyte_closeup_fig} shows a refined view of the monocyte block (clusters 3, 5, and 8) at $K=27$.
Within this block, CD14 monocytes form two major clusters: one dominated by Core CD14 monocytes and the other composed almost entirely of ISG+ CD14 monocytes, representing cells with the strongest interferon-response signature.
Interestingly, the PHM-ensemble hierarchy shows that the Core CD14 monocyte cluster merges with the CD16 monocyte cluster before being joined by the ISG+ CD14 monocyte cluster, indicating that Core CD14 monocytes overlap more with CD16 monocytes than with ISG+ CD14 monocytes in gene expression space.
This observation is consistent with the differentiation of CD16 monocytes from CD14 monocytes \cite{tak2017circulatory,patel2017fate} and suggests that the interferon-stimulated state introduces greater transcriptional differences among CD14 monocytes than those associated with the CD14-to-CD16 transition.

Using this atlas-scale dataset, we further benchmark the computational efficiency of PHM-ensemble against the Leiden clustering method used in \cite{gong2024longitudinal}.
On a computing node equipped with 3.1 GHz Intel Xeon processors, ensemble density estimation takes 1 to 2 hours of wall-clock time per replicate.
By running all 50 replicates in parallel on a SLURM-managed computing cluster, the density estimation is completed within 4 hours.
Hierarchy construction requires about 21 minutes of wall-clock time on a single computing node, including 15.5 minutes to compute the $\Delta\pmc$ matrix and 5.7 minutes for hierarchical merging.
Throughout the PHM-ensemble procedure, total memory usage across all processing units remains below 2 GB.
In comparison, our replication of the original Leiden clustering takes approximately 8.5 hours on the same system, with peak memory usage exceeding 10 GB.

\section{Discussion}

In this paper, we propose a novel computational framework for clustering large-scale datasets that addresses the challenges of fidelity, interpretability, stability, and scalability. 
Taking a probabilistic representation learning perspective, the framework first learns the underlying data-generating distribution from the observed samples and then applies the PHM algorithm to construct a cluster hierarchy from the learned distribution. 
To make this framework practical for large-scale data, we develop computational strategies for ensemble density estimation and hierarchy construction, with particular emphasis on both accuracy and stability. 
Through synthetic and real-data illustrations, we demonstrate that the resulting approach can effectively recover meaningful cluster structures in large-scale datasets while providing a flexible and interpretable multiresolution representation of the data.

Embedding and clustering are two integral components of unsupervised learning for high-dimensional and complex data. 
While considerable attention in the literature has been devoted to developing effective embedding methods, the problem of clustering in the resulting embedding space has received comparatively less attention. 
In this work, as illustrated particularly by the MNIST example, we demonstrate that the choice and performance of the clustering method can substantially influence the quality of the final output of an unsupervised learning pipeline. 
More broadly, this observation applies to many practical settings (such as spatial transcriptomics), where the benefits of a carefully learned embedding may not be fully realized without accurate and stable clustering.

Although PHM-ensemble can already be applied to embeddings generated by deep learning models, the proposed framework could be further integrated with modern deep learning architectures. 
One potential benefit of such integration is jointly optimizing embedding and density estimation objectives, which may ultimately lead to improved clustering performance. 
Extending existing deep learning approaches that incorporate GMMs \cite{jiang2017variational, dilokthanakul2016deep} represents a promising and natural direction for future research. 
Additionally, the ensemble implementation in the current framework could be adapted to better align with deep learning architectures, for example, by leveraging mechanisms such as dropout to generate ensemble diversity.

Effective hierarchy construction from a mixture density estimate is a critical step for identifying and interpreting cluster structures in our proposed framework.
The PHM algorithm provides a computationally efficient solution to this problem.
A key challenge in constructing a cluster hierarchy is the lack of a principled and computationally tractable measure of separation between clusters.
Standard statistical distances and divergences, including the KL divergence, Jensen-Shannon divergence, Bhattacharyya distance and coefficient, Hellinger distance, Wasserstein distance, and total variation distance, generally do not admit closed-form expressions for Gaussian mixtures \cite{hennig2010merging}.
Although $\pmc$, proposed in our previous work \cite{wen2026probabilistic}, still requires Monte Carlo approximation for high-dimensional data, it possesses a particularly useful pairwise decomposition property that enables efficient computation and updating throughout the agglomerative merging process.
In this work, we further demonstrate that $\pmc$ can be used to construct suitable linkage functions and naturally accommodate ensemble density estimates.
More generally, the decomposition property of $\pmc$ applies to a broad class of distributions and is not restricted to Gaussian mixtures.
This generality provides a foundation for extending the proposed framework to alternative density estimation methods in future work.

Our proposed method is also connected to {\em consensus clustering} methods in the literature \cite{monti2003consensus, fred2005combining}, which generate an ensemble of partitions of the same dataset and construct a co-association matrix to derive a final consensus partition.
While both approaches aim to improve clustering stability, they differ in the ensemble they construct.
Consensus clustering methods aggregate discrete partition labels, which requires resolving cluster correspondence across ensemble members and yields an $N \times N$ co-association matrix. 
In comparison, our proposed framework instead pools ensemble members' density estimates directly, avoiding this correspondence problem and yielding a combined representation whose size depends on the number of fitted components rather than the number of data points.
More importantly, our approach is aligned with the broader goal of multiresolution clustering: rather than terminating at a single flat partition, as in traditional consensus clustering, it provides a unified density representation that supports the subsequent construction of a full cluster hierarchy.

Finally, we do not view the proposed clustering framework as a replacement for existing partition-based or soft clustering methods. These methods are integral to the framework, providing critical inductive biases during density estimation that help capture structure in high-dimensional, multimodal spaces. We hope the proposed framework encourages the development of diverse and novel clustering algorithms to meet the urgent need for effective cluster analysis of large-scale, high-dimensional, and complex scientific data.

\section*{Software and Data Availability}

The software implementation of the PHM-ensemble procedure is available at \url{https://github.com/aturfah/PHM}.
The code and data used to generate the results in this paper are available at \url{https://github.com/aturfah/PHM_ensemble_repo}.

\section*{Acknowledgements}

This work is supported by the NIH grants R35GM138121 and R01ES033634.

\bibliographystyle{unsrt}
\bibliography{references} 

\end{document}